\documentclass[preprint]{elsarticle}
\usepackage{lineno,hyperref}
\usepackage[subrefformat=parens,labelformat=parens]{subfig}
\usepackage{amsmath}
\usepackage{soul}
\usepackage{amssymb}
\usepackage{tabularx}

\modulolinenumbers[5]

\journal{Journal of Pattern Recognition}
\usepackage{times}  
\usepackage{helvet} 
\usepackage{courier}  

\usepackage{graphicx}
\usepackage{algorithm2e}

\biboptions{authoryear}

\begin{document}

\begin{frontmatter}

\title{Joint Deep Cross-Domain Transfer Learning for Emotion Recognition}


\author[1]{Dung Nguyen\corref{cor1}} 
\cortext[cor1]{Corresponding author: 
  Tel.: +61-424-020-883;  
  }
\ead{d.nguyentien@qut.edu.au}
\author[1]{Sridha {Sridharan}} 

\author[2]{Duc Thanh {Nguyen}}

\author[1]{Simon {Denman}}

\author[3]{Son {N. Tran}} 
\author[1]{Rui {Zeng}} 
\author[1]{Clinton {Fookes}}  


\address[1]{Speech, Audio, Image and Video Technology (SAIVT) Laboratory - Queensland University of Technology}
\address[2]{School of Information Technology - Deakin University}
\address[3]{Information \& Communication Technology - University of Tasmania}
\begin{abstract}
Deep learning has been applied to achieve significant progress in emotion recognition. Despite such substantial progress, existing approaches are still hindered by insufficient training data, and the resulting models do not generalize well under mismatched conditions. To address this challenge, we propose a learning strategy which jointly transfers the knowledge learned from rich datasets to source-poor datasets. Our method is also able to learn cross-domain features which lead to improved recognition performance. To demonstrate the robustness of our proposed framework, we conducted experiments on three benchmark emotion datasets including eNTERFACE, SAVEE, and EMODB. Experimental results show that the proposed method surpassed  state-of-the-art transfer learning schemes by a significant margin.
\end{abstract}

\begin{keyword}
emotional knowledge transfer; emotion recognition; facial expression recognition; speech emotion recognition; transfer learning; cross-domain transfer; joint leaning 
\end{keyword}

\end{frontmatter}

\section{Introduction}
\label{sec1}
The well-known challenge underpinning automatic emotion recognition is the lack of sufficient labelled data to train robust models for classifying emotion. Collecting and accurately labelling emotion categories for large-scale datasets are not only costly and time consuming, but also require specific skills and knowledge \cite{8099989}. Remedies to overcome this limitation are sought to exploit the full potential of advances in deep learning techniques to improve the recognition performance on available limited datasets.

To address the problem of data scarcity in emotion classification, transfer learning has been widely adopted \cite{7956190,Ng:2015:DLE:2818346.2830593,7410698,b8e7f4bd63aa4b139eb33231552aac88}. As shown in the literature, existing methods have made use of pre-trained models that have been well trained on one dataset and fine-tune them on novel ones. Experimental results show that the knowledge captured by pre-trained models on non-target datasets can be well transferred to target ones via fine-tuning. However, these efforts have only attempted to transfer the learned emotional knowledge across various datasets within a single domain. It is shown that different domains, e.g., visual and auditory domain, provide complementary information for understanding human's emotion and thus could enrich emotion recognition models \cite{b8e7f4bd63aa4b139eb33231552aac88,JI2019231,7956190}. However, transferring knowledge across various domains, e.g., from visual to auditory domain and vice versa, is a challenging task. This kind of transfer learning also poses a greater challenge when the training/testing is performed on different datasets; considerable drop in performance is often observed due to distribution shift across datasets \cite{5995347}.


To tackle the task of transferring emotional knowledge learned across multiple domains and on multiple resource-poor datasets without suffering from distribution shift, we propose a joint deep cross-domain learning method that aims to learn cross-domain knowledge and jointly transfer the learned knowledge from rich datasets to source-poor datasets to improve the performance of emotion recognition. The transferring of features from rich datasets to source-poor datasets is performed concurrently on the source-poor datasets in an end-to-end fashion.


Our method is inspired by the face recognition method proposed in \cite{Sun:2014:DLF:2969033.2969049}, which simultaneously learns facial features using two supervisory channels: face identification and face verification. The framework enables learning compact yet discriminative features that reduce intra-person variations while enlarging inter-person differences. However, unlike \cite{Sun:2014:DLF:2969033.2969049}, in our method, pairs of training samples are taken from \textit{two different resource-poor datasets}, rather than from \textit{the same rich dataset}. \hl{Our proposed joint loss function has quite similar objectives to that proposed by [Ji et al. (2019)]. This means that the main focus of these loss functions is to minimize intra-class variations, and maximize inter-class variations to be able to learn effectively across multiple datasets. To develop a novel idea, we propose a joint deep cross-domain learning algorithm, in which three supervisory signals including two emotion classifiers and an emotion matching signal are simultaneously jointly learned in a single framework. The focus of these two emotion classifiers is to accurately detect all types of emotions
from different domains using cross-entropy loss, whereas the aim of emotion matching is to match a pair of samples to determine if they correspond to the same emotion or not relying on contrastive loss. The network proposed by [Ji et al. (2019)] composes an Intra-category Common feature representation channel (IC), an Inter-category Distinction feature representation channel (ID) for facial expression
representation, and a fusion network combining two channel features for facial expression recognition in cross databases. While the IC channel learns common features of intra-category facial expressions for
the common representation by minimizing the distance between such extracted common features, the ID channel learns Distinction features of different categories for the representation of facial expressions by maximizing distances of samples in different categories. Therefore, the main difference between our approach and one proposed by [Ji et al. (2019)] is that while the contrastive loss is exploited to match a paired sample, [Ji et al. (2019)] have utilized the distance method. In addition to this, three supervisory signals of our architecture are concurrently jointly learned, while the IC channel, the ID channel, and the ICID fusion network are sequentially learned.} As shown in our experimental results, the proposed learning framework is able to effectively learn cross-domain features and well generalize on different poorly resourced and disjoint datasets. To this end, we make the following contributions,

\begin{itemize}
    \item We investigate a relatively unexplored problem: how to effectively train an emotion recognition model on various multi-domain datasets when some or all the datasets are resource poor, in an end-to-end fashion. We address this problem by proposing a joint deep cross-domain learning method. Unlike existing works, e.g., \cite{b8e7f4bd63aa4b139eb33231552aac88,JI2019231,7956190}, our learning method is jointly performed on multiple datasets and thus is able to learn not only cross-domain features but also cross-dataset features.


    \item We investigate Group Normalization (GN) technique \cite{Wu_2018_ECCV} with different mini batches for emotion recognition. Furthermore, we report an interesting finding that GN achieves very competitive recognition accuracy with small mini batches of size of 2. We found that using small size batches did not negatively affect cross-domain transfer, yet significantly saved memory consumed in training without sacrificing the recognition accuracy.

    \item We conduct extensive experiments and develop various baselines, including face expression model, speech emotion model, and cross-domain transfer model using various off-the-shelf models to validate the proposed method.
\end{itemize}

The remainder of this paper is organized as follows: Section 2 briefly reviews related work; Section 3 describes our proposed method; Section 4 presents experiments and results; and Section 5 concludes our paper.

\section{Related work}
\label{sec:lit}
\textbf{Fine-tuning:} Transfer learning has been extensively applied in emotion recognition on visual data \cite{7956190,Ng:2015:DLE:2818346.2830593}. The most common way is to fine-tune a pre-trained model such as ResNet or AlexNet on specific visual emotion datasets. Such approach was inspired by self-taught learning \cite{He:2015:DDR:2919332.2919814}, and aims to exploit rich representations learned in a source dataset to improve the generalization in a target dataset, thereby alleviating over-fitting when training is done from scratch and with a small amount of training data. For example, in \cite{Ng:2015:DLE:2818346.2830593}, emotion recognition models were fine-tuned from models pre-trained on ImageNet. In \cite{7956190}, 3D convolutional networks, encoding both spatial and temporal information, were intially trained on a large-scale video dataset and subsequently fine-tuned on a much smaller emotion dataset to learn both audio and visual features.\\

\noindent \textbf{Cross-domain transfer:} Cross-domain transfer was investigated in \cite{b8e7f4bd63aa4b139eb33231552aac88,JI2019231}. In particular, \cite{b8e7f4bd63aa4b139eb33231552aac88} initially trained their model on the Large-scale Subtle Emotions and Mental States in the Wild database, and then transferred the learned knowledge to a traditional (non-subtle) expression dataset. Similarly, the pre-trained model in \cite{JI2019231} was learned on two different domains and fine-tuned by fusing the pre/post-trained models with a classification loss. In \cite{Albanie18a}, facial features learned from face image dataset were transferred to speech domain using distillation \cite{7780678}.\\

\noindent \textbf{Training with multiple datasets}: While exploiting cross-domain transfer was a key component of  \cite{b8e7f4bd63aa4b139eb33231552aac88,JI2019231} to overcome insufficient data, the rest of their architecture focused on addressing the domain shift between multiple datasets. Specifically, distribution alignment was adopted in  \cite{b8e7f4bd63aa4b139eb33231552aac88} to leverage tasks including subtle facial expression recognition and landmark detection on disjoint datasets. It was pointed in \cite{Zeng_2018_ECCV} that a straightforward combination of multiple datasets could not lead to any improvement of the recognition performance due to the bias and inconsistency in the annotation of the datasets and the large amount of unlabelled data. To address this issue, an Inconsistent Pseudo Annotations to Latent Truth (IPA2LT) scheme was proposed in \cite{Zeng_2018_ECCV}. In this scheme, each sample was initially assigned to more than one label by manually annotating or automatically predicting. An end-to-end LTNet was then developed to discover the latent truth from input face images and inconsistent pseudo labels.


\section{Proposed method}

\begin{figure*}[t]
\begin{center}

\includegraphics[width=1.\linewidth]{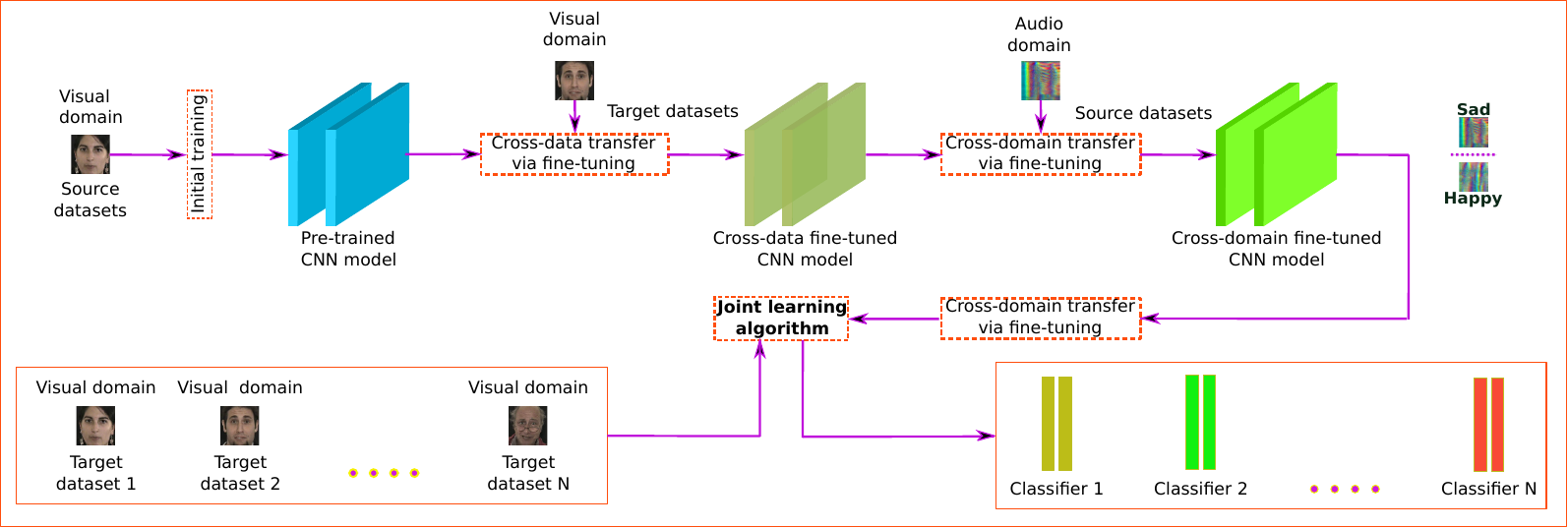}

\end{center}
   \caption{Pipeline of our proposed method. We initially pre-train our model on one visual domain then transfer the knowledge learned from the visual domain to different modalities using (continuous) fine-tuning.}
\label{methodology}
\label{fig:onecol}
\end{figure*}

\subsection{Pipeline}

Our algorithm is designed to learn emotional knowledge across visual and auditory domain, and transfer the cross-domain knowledge to multiple source-poor datasets.

Let  $D^{\mathcal{S}_v}$ be a source dataset of visual data and $D^{\mathcal{S}_a}$ be a source dataset of audio data. These source datasets are used for initial training the model, i.e., building pre-trained model. We also have $N$ visual target datasets denoted as $D^{\mathcal{T}_v}_1$, ..., $D^{\mathcal{T}_v}_N$. All target datasets are resource poor, i.e., they contain small numbers of annotated data.

We first train an initial model $P_v^p$ using the visual dataset $D^{\mathcal{S}_v}$. This initial model is also considered as pre-trained model and then fine-tuned using one of the target datasets $D^{\mathcal{T}_v}_1$, ..., $D^{\mathcal{T}_v}_N$. This step results in a cross-data fine-tuned model $F^v$. To incorporate audio features, the model $F^v$ is adapted onto the auditory dataset $D^{\mathcal{S}_a}$ and achieves a cross-domain fine-tuned model $G^{v,a}$. We finally adapt $G^{v,a}$ to the $N$ target datasets $D^{\mathcal{T}_v}_1$, ..., $D^{\mathcal{T}_v}_N$ resulting in $N$ cross-domain fine-tuned models. In order to transfer common knowledge shared by all target domains, the final $N$ cross-domain fine-tuned models are jointly trained. The pipeline of our method is illustrated in Fig. \ref{methodology}.

When training the initial model, the source dataset $D^{\mathcal{S}_v}$ is chosen as it contains the largest number of annotated samples. The choice of domain for building the pre-trained model relies on the availability of the domain data. Visual domain data is usually more accessible than audio domain data, and the model can learn better on a domain which has more annotated samples, thereby achieving a better set of parameters for further transferring/fine-tuning.

The emotional knowledge learned in the pre-trained models can be reused in cross-domain transferring steps. The reason we conduct this cross-domain transferring, i.e., transferring the learned emotional knowledge from the pre-trained model from visual domain to the auditory emotion domain prior to carrying out joint learning is because there is complementary information between visual and auditory domain. It therefore can accumulate useful emotional knowledge to the current model. The learned emotional knowledge of this model is then transferred to multiple datasets using our proposed joint learning algorithm which concurrently minimizes intra-class emotion variance and maximizes inter-class emotion variance on these resource-poor datasets. Our proposed joint learning algorithm is explained in detail in Section \ref{subsec:emotion-matching}.

\subsection{Data pre-processing}
\label{subsec:preprocessing}

\textbf{Video stream:} After extracting frames from videos, face regions are extracted using an improved Viola-Jones algorithm \cite{NGUYEN2018}. Finally, detected face regions are resized to 64$\times$64$\times$3.\\

\begin{algorithm*}[t]

\LinesNumbered
\SetAlgoLined
\KwIn{Two target domains: $D^{\mathcal{T}}_i={(x_i,l_i)}$ and $D^{\mathcal{T}}_j={(x_j,l_j)}$, parameters $\theta_{class1}$, $\theta_{class2}$, and $\theta_{match}$ transferred from our pre-trained models through continuous fine-tuning, learning rate $\eta(t)$, $t \leftarrow{0}$}

\SetAlgoLined
 \While{not converge}{
  $t \leftarrow{t+1}$ sample a number of pair of training samples $(x_i,x_j,l_i,l_j,y_{ij})$ from two mini batches which are correspondingly taken from $D^{\mathcal{T}}_i$ and $D^{\mathcal{T}}_j$;\\
  $f_i=\text{Conv}(x_i,\theta_e$) and $f_j=\text{Conv}(x_j,\theta_e$)\\
  $\Delta\theta_{class1}=\frac{\partial\text{L}(f_i,l_i,\theta_{class1})}{\partial\theta_{class1}}+\frac{\partial\text{L}(f_j,l_j,\theta_{class1})}{\partial\theta_{class1}}$\\
  $\Delta\theta_{class2}=\frac{\partial\text{L}(f_i,l_i,\theta_{class2})}{\partial\theta_{class2}}+\frac{\partial\text{L}(f_j,l_j,\theta_{class2})}{\partial\theta_{class2}}$\\
  $\Delta\theta_{match}=\lambda\cdot\frac{\partial\text{Lc}(f_i,f_j,y_{ij},\theta_{match})}{\partial\theta_{match}}$, where $y_{ij}=1$ if $l_i=l_j$, and $y_{ij}=0$ otherwise.\\

  $f_i=\frac{\partial\text{L}(f_i,l_i,\theta_{class1})}{\partial f_i}+\frac{\partial\text{L}(f_i,l_i,\theta_{class2})}{\partial f_i}+\lambda\cdot\frac{\partial\text{Lc}(f_i,f_j,y_{ij},\theta_{match})}{\partial f_i}$\\

  $f_j=\frac{\partial\text{L}(f_j,l_j,\theta_{class1})}{\partial f_j}+\frac{\partial\text{L}(f_j,l_j,\theta_{class2})}{\partial f_j}+\lambda\cdot\frac{\partial\text{Lc}(f_i,f_j,y_{ij},\theta_{match})}{\partial f_j}$\\

  $\Delta\theta_e=\Delta f_i\cdot\frac{\partial\text{Conv}(x_i,\theta_e)}{\partial\theta_e}+\Delta f_{j}\cdot\frac{\partial\text{Conv}(x_j,\theta_e)}{\partial\theta_e}$\\
  update $\theta_{match}=\theta_{match}-\eta(t)\cdot\Delta\theta_{match}$, $\theta_{class1}=\theta_{class1}-\eta(t)\cdot\Delta\theta_{class1}$, $\theta_{class2}=\theta_{class2}-\eta(t)\cdot\Delta\theta_{class2}$, and $\theta_e=\theta_e-\eta(t)\cdot\Delta\theta_e$\\
 }

 \caption{Our joint learning algorithm}
 \KwOut{$\theta_e$}
 \label{algorithms}
\end{algorithm*}

\noindent \textbf{Audio Stream:}
\cite{8085174} extracted three channels of log Mel-spectrograms from segments over all utterances. They fine-tuned a pre-trained AlexNet model on such features, achieving significantly better performance than with hand-crafted features. Inspired by this, we also extract three channel features as follows. First, Mel-spectrogram segments with size 64 $\times$ 64 $\times$ 3 ($F$ = 64, $T$ = 64, $C$ = 3) are generated from 1-D speech signals, with $F$, $T$, and $C$ denoting the number of Mel-filter banks, the segment length corresponding to the frame number in a context window, and the number of channels of the Mel-spectrogram, respectively. For cross-domain transfer we select values of $F$ and $T$ to match the pre-processed video, while the 3 channels of the Mel-spectrograms are the static, delta, and delta-delta coefficients, similar to \cite{8085174}. Next, we convert 64 Mel-filter banks from 20 to 8000 Hz into a log Mel-spectrogram using a 25 ms Hamming window with a 10ms overlap for an utterance. A context window of 64 frames (length 10 ms $\times$ 63 + 25 ms = 655 ms) is then applied to the whole log Mel-spectrogram to extract the static 2-D Mel-spectrogram segments (64 $\times$\ 64) with an overlap size of 30 frames \cite{8085174}.

\subsection{Joint learning}
\label{subsec:emotion-matching}
To accumulate the emotional knowledge from visual domain and auditory domain which is subsequently transferred and re-used as initial knowledge for our joint learning on multiple visual domains, we initially pre-train our convolutional neural network on $D^{\mathcal{S}_v}$, and then transfer the learned emotional knowledge from our pre-trained visual emotion model to $D^{\mathcal{T}_v}_1$, ..., $D^{\mathcal{T}_v}_N$ for recognition of visual emotion, and then transfer it to $D^{\mathcal{S}_a}$ for extraction of speech feature using a continuous fine-tuning technique proposed by \cite{8099989} (see Table \ref{list_of_experiment}). During pre-training and (continuous) fine-tuning, these models are learned by maximizing the posterior probability of the ground-truth, focusing on separating features from different classes \cite{Wang_2018_CVPR}. Given an input feature vector $x$ with its corresponding ground-truth label $c$, the cross-entropy loss is formulated as follows:
\begin{equation}\label{Emo_class}
 L(f,c,\theta_{class}) = -\sum_{i=1}^{C}p_i\text{log}\hat{p}_i=-\text{log}\hat{p}_c,
\end{equation}

Where $f = Conv(x, \theta_e)$, $Conv(\cdot)$ is defined by our proposed convolutional neural network (CNN) and is then used as the feature extraction function, $x$ is the input face image or input speech segment for cross-domain transfer, and $\theta_e$ denotes our CNN parameters to be learned which are randomly initialized when pre-training and are transferred from our pre-trained models when fine-tuning; $c$ and $\theta_{class}$ denote the target class and the softmax layer parameters, respectively; $p_i$ is the target probability distribution, $p_i$ = 0 for all $i$ except $p_c=1$ for the target class $c$, $\hat{p}_i$ is the predicted probability distribution.


The set of accumulated parameters of our model fine-tuning on the final domain is now transferred and reused as initial knowledge for our joint learning on two different visual domains by simultaneously optimizing two cross-entropy losses of E.q.(~\ref{Emo_class}) and a contrastive loss, shown in  E.q.(~\ref{Emo_match}), as originally proposed by \cite{1640964} for dimensionality reduction (see Fig. \ref{methodology}).

The contrastive loss is calculated as follows,
\begin{equation} 
\label{Emo_match}
  L_c=
  \begin{cases}
      = \frac{1}{2} \|f_i-f_j\|_2^2& \text{if $y_{ij}$ = 1}\\
      \frac{1}{2}max(0,m-\|f_i-f_j\|_2^2) & \text{if $y_{ij}$ = 0}
    \end{cases},
\end{equation}
where $f_i$ = $Conv(x_i,\theta_e)$, $f_j$ = $Conv(x_j,\theta_e)$, $x_i$ and $x_j$ are taken from $D^{\mathcal{T}_v}_i$ and $D^{\mathcal{T}_v}_j $, and $m$ is the predefined margin. If $x_i$ and $x_j$ are from the same emotion, then $y_{ij}$ = 1. In this case, E.q.~(\ref{Emo_match}) minimizes the $\mathcal{L}2$ distance between the two feature vectors: $f_i$ and $f_j$. If $x_i$ and $x_j$ are from different emotions, then $y_{ij}$ = 0. A mathematical framework similar to ours is proposed in \cite{Nagrani18c,Sun:2014:DLF:2969033.2969049,Nagrani18a}. However, $x_i$ and $x_j$ are a pair of face images always taken from one domain and $\theta_e$ is randomly initialized in \cite{Sun:2014:DLF:2969033.2969049}. In contrast, in our joint learning algorithm, ($x_i$, $x_j$) is a pair of samples taken from $D^{\mathcal{T}_v}_i$ and $D^{\mathcal{T}_v}_j$, and $\theta_e$ is transferred from a set of learned parameters of our model fine-tuning on the final auditory domain. Likewise, only a cross-entropy loss function in E.q.(~\ref{Emo_class}) was optimized in \cite{Nagrani18a} and only a contrastive loss function was optimized by \cite{Nagrani18c}, whereas all supervisory signals consisting of two cross-entropy loss functions ($L_1$ and $L_2$) (E.q.(~\ref{Emo_class})) and a contrastive loss function ($L_c$) (E.q.(~\ref{Emo_match})) are jointly optimized in our framework.

Therefore, the training loss function $L_{joint}$ for our joint learning algorithm is defined as follows,
\begin{equation}
L_{joint}=\lambda{_1}L_{1}+\lambda{_2}L_{2}+\lambda{_3}L_{c},
\end{equation}
where $\lambda{_1}$, $\lambda{_2}$, and $\lambda{_3}$ are hyper-parameters.

What is the motivation for such joint learning? While the aim of classification signals is to classify each sample from multiple domains into different types of emotions by maximizing inter-class variations, the objective of the matching signal is to predict whether a pair of samples belongs to the same emotion by reducing intra-class variations.

\newcolumntype{y}{>{\hsize=8.7cm}X}
\newcolumntype{Y}{>{\hsize=2.8cm}X}
\newcolumntype{K}{>{\hsize=2.5cm\centering\arraybackslash}X}
\newcolumntype{B}{>{\hsize=4.5cm\centering\arraybackslash}X}
\newcolumntype{A}{>{\hsize=14.2cm\centering\arraybackslash}X}

\begin{table*}[t]
\begin{center}
\caption{All sets of experiments of our joint deep cross-domain learning framework conducted on visual eNTERFACE, visual SAVEE, audio SAVEE, and audio EMODB}

\begin{tabularx}{\textwidth}{Y|y}

 \textbf{Set of experiment}& \textbf{Description}\\
\hline

V$_{\text{eNTER}}\textunderscore$Model&Pre-training on visual eNTERFACE dataset\\
\hline
V$_{\text{SAV}}\textunderscore$Model (Fine-tuned)& Fine-tuning fully-connected layers of the pre-trained V$_{\text{eNTER}}\textunderscore$Model on visual SAVEE dataset\\
\hline
A$_{\text{SAV}}\textunderscore$Model &Pre-training on audio SAVEE\\

\hline
A$_{\text{SAV}}\textunderscore$Model (Fine-tuned) &Fine-tuning V$_{\text{SAV}}\textunderscore$Model (Fine-tuned) on audio SAVEE\\

\hline
A$_{\text{EMO}}\textunderscore$Model&Pre-training on audio EMODB\\

\hline
A$_{\text{EMO}}\textunderscore$Model (Fine-tuned) &Fine-tuning A$_{\text{SAV}}\textunderscore$Model (Fine-tuned) on audio EMODB\\
\hline

V$_\text{eNTER+SAV}\textunderscore$Model&Pre-training on visual eNTERFACE and visual SAVEE\\

\hline

Our joint learning algorithm&Our joint learning using continuous fine-tuning A$_{\text{EMO}}\textunderscore$Model (Fine-tuned) on visual SAVEE and visual eNTERFACE by optimizing our joint loss function
\end{tabularx}
\label{list_of_experiment}

\end{center}
\end{table*}

Our goal is to learn the parameters $\theta_e$  in the feature extraction function $Conv(\cdot)$, while $\theta_{match}$, $\theta_{class1}$, and $\theta_{class2}$ are parameters introduced to propagate two emotion classification signals and an emotion matching signal. The parameters are updated by stochastic gradient descent. The emotion classification and emotion matching gradients are weighted by a hyperparameter. Our joint learning algorithm is summarized in Algorithm \ref{algorithms}. During testing, $\theta_e$ is used for feature extraction.

How to generate paired samples for our joint learning algorithm? Paired samples are input for the contrastive loss, and are usually generated from two datasets. However, this approach may create a huge number of pairs, consequently slowing fine-tuning. We instead propose to generate paired samples from each mini batch. We suppose that two mini batches of size $K$ from two corresponding datasets are fed into our framework at every iteration for joint learning, and a total of $K^2$ pairs of samples are generated. Our model can learn better if all pairs of samples are generated from each pair of mini batches at each iteration. Therefore, it is required for our joint learning algorithm to use a small mini batch since in addition to saving memory, all pairs of samples can be generated without significantly increasing the number of pairs.

Unfortunately, batch normalization (BN) \cite{pmlr-v37-ioffe15} performs well only with a large mini batch, while with a small mini batch it inaccurately estimates the mini batch statistics, increasing model error \cite{Wu_2018_ECCV}. In contrast, our experimental results further confirm that Group Normalization (GN) \cite{Wu_2018_ECCV} achieves similar performance with batch sizes from 2 to 512. Following this finding, we exploit GN with a mini batch-size of 2 for our approach. We note that GN with a mini batch-size of 2 does not affect a natural transfer process since GN is independently computed along mini batches, while with a mini batch-size of 2 BN becomes a linear layer $y=\frac{\gamma}{\sigma}(x-\mu+\beta)$, where $\mu$ and $\sigma$ were previously computed from the pre-trained model and frozen \cite{He:2015:DDR:2919332.2919814}. These are the main reasons why we initially investigate different mini batch-sizes for GN before carrying out the extensive cross-domain transfer and joint learning experiments.

Our general joint deep cross-domain learning framework was formulated for $N$ visual and $M$ audio source databases and $J$ visual and K audio target databases. To demonstrate and evaluate the performance our method we use a small number for $N$, $M$, $J$, and $K$. Specifically $N=1$, $M=1$, $J=1$, and $K=3$. For our joint learning algorithm, our system was formulated for learning on 2 different visual datasets.

\section{Experiments \& Results}
\label{experiment}
\textbf{Dataset Details:} The eNTERFACE dataset \cite{1623803} is an audio-visual dataset which has 44 subjects and includes a total of 1293 video sequences in which the proportion of sequences corresponding to women and men are 23\% and 77\%, respectively. Subjects were asked to express 6 discrete emotions: anger, disgust, fear, happiness, sadness, and surprise \cite{1623803}.

The SAVEE dataset \cite{HaqJackson_AVSP09} is an audio-visual dataset which was recorded by higher degree researchers (aged from 27 to 31 years) at the University of Surrey, and four native male British speakers. All of them were also required to speak and express seven discrete emotions: anger, disgust, fear, happiness, sadness, surprise, and neutral. The dataset contains 120 utterances per speaker \cite{HaqJackson_AVSP09}.

The EMO-DB dataset \cite{Burkhardt05adatabase} is an acted speech corpus containing 535 emotional utterances with seven different acted emotions listed: anger, disgust, fear, happiness, sadness, surprise, and neutral. These emotions were stimulated by five male and five female professional native German-speaking actors, generating five long and five short German utterances used in daily communication. These actors were asked to read predefined sentences in the targeted emotions.\\

\noindent \textbf{Network architecture:} The network we use for all stages is VGG-16. The network includes four convolutional layers with 64, 128, 256, and 512 3$\times$3 filters and stride 1, respectively, the output of each convolutional layer is activated using leaky rectified linear units (LReLU) \cite{Maas13rectifiernonlinearities} before being normalized by Group Normalization \cite{Wu_2018_ECCV}, afterwhich the network has  four fully-connected layers with hidden units of dimensionality 512, 128, 32, and 6, respectively. A LReLU activation function is also exploited after each fully-connected layer.\\

\noindent \textbf{Implementation details:} For pre-training, we apply the variance scaling initialiser of \cite{8652371} that has been recently proposed for network weight initialization for all convolutions. Group Normalization \cite{Wu_2018_ECCV} and  Dropout (0.5) are exploited in all convolutional layers. We train our model for 100,000 iterations with a learning rate of 10e-5. For cross-domain transfer using (continuous) fine-tuning, we fine-tune only fully-connected layers when transferring between visual domains, and continuously fine-tune all layers when transferring from visual to audio domains. For joint learning using continuous fine-tuning, we set $\lambda{_1}=1$, $\lambda{_2}=1$, and $\lambda{_3}=0.01$, and reduce the learning rate to 10e-6. The models are trained for 20,000 iterations and the mini batch-size is fixed at 2. In all sets of our experiments, we apply k-fold cross-validation, the original training data is randomly divided into k equal parts. Of the k-parts, one of them is fixed as the validation data for testing the model, and the other k-1 parts are used as training data. The cross-validation process is then repeated 5 times. We only focus on recognizing 6 emotions: anger, disgust, fear, happiness, sadness, and surprise.\\

\noindent \textbf{Video emotion recognition results:} Experimental results of video emotion recognition models: V$\textunderscore$eNTER$\textunderscore$Model, which is pre-trained on video eNTERFACE, and V$\textunderscore$SAV$\textunderscore$Model (Fine-tuned), which fine-tunes only fully-connected layers of the pre-trained V$\textunderscore$eNTER$\textunderscore$Model on video SAVEE are illustrated in Table \ref{Resutls of video eNTER} and Table \ref{Resutls of video SAVEE}, respectively. As shown in Table \ref{Resutls of video eNTER} and Table \ref{Resutls of video SAVEE}, the V$\textunderscore$SAV$\textunderscore$Model (Fine-tuned) and the V$\textunderscore$eNTER$\textunderscore$Model achieve state-of-the-art performances with approximately 99\% and 93\% recognition accuracy for the within-corpus scenario on visual SAVEE and visual eNTERFACE, accordingly. However, these performances decrease significantly when evaluating these models in regard to a cross-corpus scenario as illustrated in Table \ref{our proposed system for video emotion recognition}. For example, the V$\textunderscore$SAV$\textunderscore$Model (Fine-tuned) shows a significant drop of 79\% in emotion recognition accuracy when evaluated on visual eNTERFACE. Similarly, V$\textunderscore$eNTER$\textunderscore$Model achieves only 35\% emotion recognition accuracy when testing on visual SAVEE (see Table \ref{our proposed system for video emotion recognition}). This significant drop is attributed to the distribution shift across datasets.\\

\begin{table}[t]

\begin{center}
\caption{Results of our proposed model evaluated on the video eNTERFACE for visual emotion recognition}
\label{Resutls of video eNTER}
\begin{tabular}{l||cc}
\hline
\textbf{Method} &    \textbf{Acc}\\
\hline\hline
CNN Model \cite{7945502} &0.64\\

\hline
Fine-Tuned Pre-trained C3D \cite{7956190}& 0.54\\
\hline
Video C3D + DBN \cite{7926723}&0.83\\
\hline
\textbf{V$\textunderscore$eNTER$\textunderscore$Model (Fine-tuned, 1-part)}&\textbf{0.92}\\
\hline
\textbf{V$\textunderscore$eNTER$\textunderscore$Model (BS=2)}&\textbf{0.93}\\
\hline
\textbf{V$\textunderscore$eNTER$\textunderscore$Model (BS=512)}&\textbf{0.93}\\
\hline
\end{tabular}
\end{center}
\end{table}

\begin{table}[t]
\begin{center}
\caption{Results of our proposed model evaluated on the audio SAVEE for speech emotion recognition}

\label{Resutls of audio SAVEE}
\begin{tabular}{l||c}
\hline
\textbf{Method} &\textbf{Acc}\\
\hline\hline
SVM \cite{7945502} &0.49\\
\hline
SVM-PCA \cite{7945502} &0.43\\
\hline
RF \cite{7945502} &0.56\\
\hline
RF-PCA \cite{7945502} &0.53\\
\hline
AlexNet \cite{8085174} &\textbf{0.69}\\

\hline
\textbf{A$\textunderscore$SAV$\textunderscore$Model} &0.59\\
\hline
\textbf{A$\textunderscore$SAV$\textunderscore$Mode (Fine-tuned)}&0.62\\
\hline

\end{tabular}
\end{center}

\end{table}
\noindent \textbf{Cross-domain transfer results:} We continuously fine-tune all layers of our V$\textunderscore$SAV$\textunderscore$Model (Fine-tuned) on audio SAVEE (see Table \ref{Resutls of audio SAVEE}), and then on audio EMODB (see Table \ref{Resutls of EMODB}) in an end-to-end fashion as described in detail in Table \ref{list_of_experiment}. As shown in Table \ref{Resutls of audio SAVEE}, the A$\textunderscore$SAV$\textunderscore$Model (Fine-tuned) achieves very promising recognition accuracy (62\%) standing in second place, which is around 3\% higher than the accuracy obtained by the A$\textunderscore$SAV$\textunderscore$Model which is trained from scratch. Similarly, the A$\textunderscore$EMO$\textunderscore$Model (Fine-tuned) demonstrates best performance (89\%) in comparison with other state-of-the-art speech emotion recognition models and is significantly better than that of the A$\textunderscore$EMO$\textunderscore$Model (67\%), as shown in Table~\ref{Resutls of EMODB}.\\

\begin{table}[t]

\begin{center}
\caption{Results of our proposed model evaluated on the video SAVEE for visual emotion recognition}
\label{Resutls of video SAVEE}
\begin{tabular}{l||cc}
\hline
\textbf{Method} &    \textbf{Acc}\\
\hline\hline

SVM-PCA \cite{7945502}& 0.52\\
\hline
RF \cite{7945502} & 0.56\\
\hline
CNN Model \cite{7945502}&0.97\\

\hline
\textbf{V$\textunderscore$SAV$\textunderscore$Model (Fine-tuned, 1-part)}&\textbf{0.96}\\
\hline
\textbf{V$\textunderscore$SAV$\textunderscore$Model (Fine-tuned, BS=2)}&\textbf{0.99}\\
\hline
\textbf{V$\textunderscore$SAV$\textunderscore$Model (Fine-tuned, BS=512)}&\textbf{0.99}\\

\hline
\end{tabular}
\end{center}
\end{table}

\begin{table}[t]
\begin{center}
\caption{Results of our proposed model evaluated on the audio EMODB for speech emotion recognition}
\label{Resutls of EMODB}
\begin{tabular}{l||c}
\hline
\textbf{Method} & \textbf{War}\\
\hline\hline

ComParE set \citeyear{7160715} &0.86\\
\hline

Alexnet-DTPM \citeyear{8085174}&0.76\\
\hline

DCNN-DTPM \citeyear{8085174}&0.84\\
\hline
Fine-Tuned Alexnet-Average \citeyear{8085174}&0.83\\
\hline
Fine-Tuned Alexnet-DTPM \citeyear{8085174}&0.87\\
\hline
\textbf{A$\textunderscore$EMO$\textunderscore$Model} &0.67\\
\hline

\textbf{A$\textunderscore$EMO$\textunderscore$Model (Fine-tuned)} &\textbf{0.89}\\
\hline

\end{tabular}
\end{center}

\end{table}
\begin{table}[t]
\begin{center}
\caption{The training error of our cross-domain transfer models in comparison with other model errors of cross-domain transfer models using off-the-shelf models when fine-tuned on video SAVEE and video eNTERFACE}
 \label{baseline results}
\begin{tabular}{l||c|c}
\hline
\textbf{Method} & Error (SAV)&Error (eNTER)  \\ \hline \hline
Cifar \citeyear{6751366} & 1.773 &1.3263\\
Inception$\_$v1 \citeyear{7298594} & 1.812 &- \\
Mobilenet$\_$v2 \citeyear{DBLP:journals/corr/HowardZCKWWAA17} & 1.796 &-\\
Resnet$\_$v1$\_$152 \citeyear{7780459}&1.886&-\\
Resnet$\_$v2$\_$50 \citeyear{He2016}&1.787&-\\
Vgg$\_$19 \citeyear{DBLP:journals/corr/SimonyanZ14a}&1.792 &1.56\\

Transfer learning \citeyear{8545411}&0.44&0.91\\
\hline

\textbf{Our proposed model} &\textbf{0.041}&\textbf{0.136}\\
\hline

 \end{tabular}
 \end{center}

 \end{table}

To further show the effectiveness of our proposed model with respect to handling insufficient data, we conduct two additional sets of experiments as follows: 1) We fine-tune fully-connected layers of our V$\textunderscore$SAV$\textunderscore$Model (Fine-tuned) on 1-part training data of video eNTERFACE (see Table \ref{Resutls of video eNTER}), and 2) We fine-tune fully-connected layers of our pre-trained V$\textunderscore$eNTER$\textunderscore$Model on 1-part training data of video SAVEE (see the Table \ref{Resutls of video SAVEE}). It is noted that 4-parts of training data are used for pre-training and fine-tuning in previous sets of experiments. As seen in the Table \ref{Resutls of video eNTER} and the Table \ref{Resutls of video SAVEE}, despite being fine-tuned on only 1-part of the training data, the V$\textunderscore$eNTER$\textunderscore$Model (Fine-tuned, 1-part) and V$\textunderscore$SAV$\textunderscore$Model (Fine-tuned, 1-part) achieve very competitive recognition accuracies (92\% and 96\%) on video eNTERFACE and video SAVEE, respectively in comparison with those when these models learn on 4-parts training data.

We re-implement other cross-domain transfer approaches using off-the-shelf models including all versions of ResNet \cite{7780459}, of Inception \cite{7298594}, of MobileNet \cite{DBLP:journals/corr/HowardZCKWWAA17}, and we also re-implement meta transfer learning \cite{8545411} to further compare and thoroughly demonstrate the efficiency of our proposed model. As shown in the Table \ref{baseline results}, our model achieves the lowest errors when fine-tuned on video eNTERFACE and on video SAVEE, accordingly. Moreover, while ResNet \cite{7780459}, Inception \cite{7298594}, and MobileNet \cite{DBLP:journals/corr/HowardZCKWWAA17} over-fit (shown as ~`-' in the Table \ref{baseline results}) when fine-tuned on video eNTERFACE, our model performs very well, achieving model error of 0.136. When fine-tuned on video SAVEE, these off-the-shelf models achieve quite competitive model errors (approximately 1.8) compared to those obtained by Cifar \cite{6751366} and Vgg$\_$19 \cite{DBLP:journals/corr/SimonyanZ14a}, which are however significantly higher than our model error (0.041).

Our cross-domain systems outperforms significantly the recent state-of-the-art emotion recognition systems relying on the transfer learning method which fine-tunes off-the-shelf/pre-trained models. This can be explained that the representational structures that are unrelated to emotion are still remained in off-the-shelf/pre-trained models and the extracted features are usually vulnerable to identity variations, leading to degrading the performance of these emotion recognition systems fine-tuning off-the-shelf/pre-trained models on the  emotion dataset.

As analysed earlier, our models achieved from the cross-domain transfer stage still poorly perform on new domains However, our objective of this stage is to accumulatively learn emotional knowledge from visual domains and from auditory domains which is then transferred to our joint learning on multiple visual domains.\\

\noindent \textbf{Joint learning results:} \hl{As explained earlier, the transfer learning task poses a greater challenge when the training/testing is performed on different datasets; significant drop in performance is often observed owning to distribution shift across datasets. To address the transferring emotional knowledge task learned across multiple domains and on multiple resource-poor datasets without suffering from distribution shift, we have proposed a joint deep cross-domain learning approach, focusing on learning cross-domain knowledge. To demonstrate the effectiveness of our joint deep cross-domain learning method, we conduct three baseline systems: 1) Our network architecture is learned only on eNTERFACE training set using classification loss and evaluated on eNTERFACE testing set and SAVEE testing set, 2) Our network architecture is learned only on SAVEE training set using classification loss and evaluated on SAVEE testing set and eNTERFACE testing set, and 3) Our network architecture is trained on both eNTERFACE training set and SAVEE training set using classification loss and evaluated on eNTERFACE testing set and SAVEE testing set (see Table 1 for further description of each experiment and notation). Whereas joint learning experiments are to train our network architecture on both eNTERFACE training set and SAVEE training set using our proposed joint loss function and evaluate on eNTERFACE testing set and SAVEE testing set (see Table 1) }. As shown in Table \ref{our proposed system for video emotion recognition}, the V\textunderscore SAV\textunderscore eNTER\textunderscore Model does not show a significant improvement in recognition accuracy, despite learning with multiple datasets enlarged by simply combining visual eNTERFACE and visual SAVEE. This model achieves only 64\% recognition accuracy which is the same as that in V\textunderscore eNTER\textunderscore Model and 4\% higher than that achieved by the V\textunderscore SAV\textunderscore Model (Fine-tuned) while validating on both video eNTERFACE and video SAVEE. Although, the V\_ SAV\textunderscore eNTER\textunderscore Model is learned by jointly optimizing two cross-entropy losses, this model still suffers from a distribution shift across datasets. In contrast, as can be seen in Table \ref{our proposed system for video emotion recognition}, the performance of our proposed model is vastly improved when learning with our proposed joint learning algorithm as described in detail in Table \ref{algorithms}, which simultaneously optimizes two cross-entropy losses and a contrastive loss. Through this approach, our proposed joint learning model achieves 66\% and 94\% emotion recognition accuracy tested on visual SAVEE and visual eNTERFACE accordingly corresponding an average accuracy of \textbf{80\%}, which shows the best performance compared to our baseline systems (V\textunderscore SAV\textunderscore Model (Fine-tuned), V\textunderscore eNTER\textunderscore Model, and V\textunderscore SAV\textunderscore eNTER\textunderscore Model) (see Table~\ref{our proposed system for video emotion recognition}).\\

\begin{table}[t]
\begin{center}
\caption{Results of our proposed joint leraning model validated on eNTERFACE and SAVEE compared with our baselines including V\textunderscore SAV\textunderscore Model (Fine-tuned), V\textunderscore eNTER\textunderscore Model, and  V\textunderscore SAV\textunderscore eNTER\textunderscore Model for visual emotion recognition}
\label{our proposed system for video emotion recognition}
\begin{tabular}{l||c|c|cc}
\hline
 \textbf{Model}&SAV &eNTER &\textbf{Avg.}\\
\hline\hline
V\textunderscore SAV\textunderscore Model (Fine-tuned)&0.99&0.20&0.60\\
\hline
V\textunderscore eNTER$\_$Model&0.35&0.93&0.64\\
\hline
V\textunderscore SAV\textunderscore eNTER\textunderscore Model&0.34&0.95&0.64\\
\hline
\hline
\textbf{\textbf{Avg.}
}&0.56&0.69&0.62\\
\hline
\textbf{\textbf{Our joint learning model}
}&0.66&0.94&\textbf{0.80}\\
\hline
\end{tabular}
\end{center}
\end{table}

Thus this demonstrates that our proposed model can generalize well across multiple datasets, hence successfully tackling distribution shift across datasets. Moreover, it is worth noting that our proposed joint learning method is fine-tuned on our pre-trained model with a small mini batch-size of 2 rather than training from scratch, therefore, saving a huge amount of time and memory resources for training without sacrificing emotion recognition accuracy. The choice of a mini batch-size of 2 is due to the following motivation: the V\textunderscore SAV\textunderscore Model (Fine-tuned) and  the V\textunderscore eNTER\textunderscore Model using a mini batch-size of 2 obtain very competitive performance compared to a mini batch-size of 512. Thus, with a small mini batch-size, we still can carry out extensive experiments with a limited memory/GPU and the number of paired samples for the contrastive loss is only 4 for each mini batch-size instead of $K^2 (K>2)$.  The reason we are unable to compare our proposed joint learning algorithm with the state-of-the-art models is that as far as we know this is the first research in addressing a training problem with multiple insufficient datasets where there is a distribution shift across datasets. As such, we only compare our joint learning results with those of our baseline models.

\section{Conclusion}

In this paper, we have developed a framework which is able to learn well with multiple poorly resourced and disjoint emotion datasets by simultaneously minimizing intra-class variance and maximizing inter-class variance. By integrating the cross-domain transfer using a continuous fine-tuning strategy, our proposed framework has successfully transferred emotional learned knowledge between modalities such as from one visual domain to another visual domain, from the visual domain to the audio domain, and then to multiple domains. To the best of our knowledge, our joint learning algorithm is the first study aimed at resolving the training problem with multiple poorly resourced emotion datasets. To validate the effectiveness of our proposed framework in learning, extensive experiments have been conducted in visual and speech emotion recognition and demonstrate that our framework performs significantly better than other state-of-the-art approaches involving three emotion datasets: eNTERFACE, SAVEE, and EMODB.
\section{Acknowledgement}
This research was supported by an Australian Research Council (ARC) Discovery grant DP140100793.

\bibliographystyle{model2-names}\biboptions{authoryear}
\bibliography{refs}
\end{document}